Published as a paper at AJIC

# A Novel Method of Fuzzy Topic Modeling based on Transformer Processing

Tseng, Ching-Hsun
*Institue of Technology Management*
National Chiao Tung University
Hinschu, Taiwan
hank131415go61@gmail.com

Lee, Shin-Jye
*Institue of Technology Management*
National Chiao Tung University
Hinschu, Taiwan
camhero@gmail.com

Cheng, Po-Wei
*SparkAmplify*
Taipei, Taiwan
kiollpt@gmail.com

Lee, Chien
*SparkAmplify*
Taipei, Taiwan
chien@sparkamplify.com

Hung, Chih-Chieh
*Department of Information Management*
*National Chung Hsing University*
smalloshin@gmail.com

***Abstract*— Topic modeling is admittedly a convenient way to monitor markets trend. Conventionally, Latent Dirichlet Allocation, LDA, is considered a must-do model to gain this type of information. By given the merit of deducing keyword with token conditional probability in LDA, we can know the most possible or essential topic. However, the results are not intuitive because the given topics cannot wholly fit human knowledge. LDA offers the first possible relevant keywords, which also brings out another problem of whether the connection is reliable based on the statistic possibility. It is also hard to decide the topic number manually in advance. As the booming trend of using fuzzy membership to cluster and using transformers to embed words, this work presents the fuzzy topic modeling based on soft clustering and document embedding from state-of-the-art transformer-based model. In our practical application in a press release monitoring, the fuzzy topic modeling gives a more natural result than the traditional output from LDA.**

## I. Introduction

In natural language processing, the typical way is to collect all vocabularies into a dictionary and then tokenize each word into a dictionary. The typical way to extract the topics among these articles is using Latent Dirichlet Allocation, LDA. By calculating the number of each word, LDA offers the contribution property based on word-to-word conditional probability. Although this approach is admittedly straightforward in a small amount of dataset, the output might be unintuitive when we face an enormous dataset because a significant number of words could share the conditional probability for each word. Also, it is the norm to suffer an excellent computing cost in building an enormous dictionary. Then, in such a sparse vector towards one word, the chances of experiencing the curse of dimensionality is notoriously high. Even worse, in a real world task, it is necessary to update the dictionary and run the LDA from scratch. Thus, finding a better way to implement topic analysis is expected to gain a better result without setting the topic number in advance.

In this dilemma, this work applies a novel model to extract a more meaningful vector towards word or document, to begin with, instead of relying on the possibility. Because of the booming development of the transformers, such as bidirectional encoder representation from the transformer, BERT(Devlin, Chang et al. 2018), and the merit of attention effect, vector not only represents the meaning of the word itself but also contains the relationship towards every word in a document. By extracting output from the hidden layer of the transformers, each word embedding is shown in multiple continuous figures, which can be viewed as a relative fuzzy index towards each direction because of the multi-head attention in a transformer. Whereas training a transformer still needs to build a token dictionary, it is, as a result, more efficient and accessible to use a pre-trained weight of BERT, which is released by Google. Most importantly, one of the significant advantages of using a transformer is because of the merit of calculating the dot-product in multi-head attention we can speed up the process by the parallelizing workflow in GPUs.

On the other hand, the pooling of mean to reshape each document embedding in the same dimension is applied in this work. It is hard to visualize the embedding in such a high dimension, so it is necessary to make a dimension reduction, projecting the vector into a low dimension space. Therefore, visualizing the relative location between documents for clustering similar data points in 2-D space becomes possible since it is not wise to cluster data in high dimensional space directly. Typically, the standard way is using linear discriminant analysis and principal component analysis. However, most dimensional reduction methods assume the data is either linearly separable or following a normal distribution. Also, although the methods above can successfully reduce the vector's dimension, the data location cannot entirely reflect the actual distance between the vector to vector in the original dimension.

In contrast, t-distributed stochastic neighbor embedding can handle these concerns well because it experiences a complicated step to project data step-by-step. It calculates each data's original distribution and clusters the similar data in low dimensional space according to the t-distribution. Our experiments reveal that t-SNE usually gives a better result than PCA and LDA as t-SNE helps cluster data based on the distribution in advance instead of merely projecting data.

After a series of preprocessing, the goal of this work is to find a better topic among a large amount articles. Instead of using conditional probability to indicate the possible topic, clustering the document embedding of each article in low dimension without setting the wanted topic number in advance is applied. Among a range of clustering models, including K-means, KNN, and DBSCAN, HDBSCAN outshines all because of not only using density to cluster data but also utilizing a hierarchy method to find the reasonable group. Instead of assuming the data is in the round group and grouping data in a dense area, HDBSCAN can cluster data in various shape and also pay more attention to handling outlier and less dense data point. Nonetheless, HDBSCAN cannot find the representative of the group, which means it is not easy to know which data is the centre of this group or which one belongs to this cluster. In order to fix the problem, applying a fuzzy membership value to data is applied in this work. With this value, it assumes the data with high membership value indicates the data is near the centre of the group or strongly represents the cluster. The experiments show extracting top data with sorting membership value not only helps to know the representative but also helps to not wrongly pick up the non-representative data on the edge of a group.

In light of the concerns above and the current trend in NLP tasks, what makes this work shine is ushering transformer method into topic modeling. Also, for a better topic representing performance, this work uses the fuzzy membership value as the determined standard. The detail about the proposed work is revealed in the following section. Apart from the simple introduction of the motivation and background of our method in this section, the related works are shown in section II which contains the present method regarding document embedding, transformer, dimension reduction, clustering in a hard and soft way. In section III, the used algorithm of the proposed model and fuzzy membership value are presented. Then, part IV includes experiment results toward different model setting and the final output in a commercial application. Lastly, conclusion and potential about the fuzzy topic model are presented in part V. The structure of the proposed model is shown below:

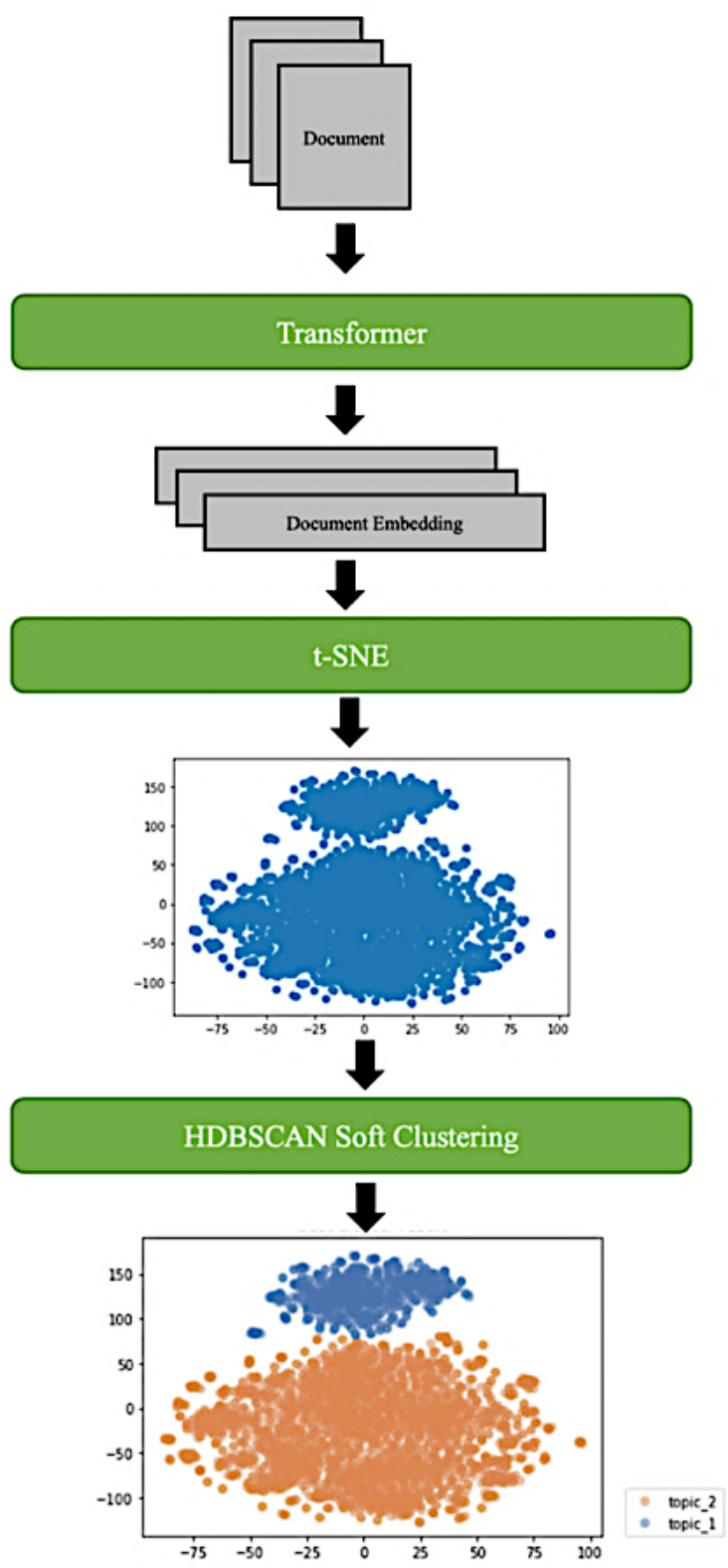

Figure 1. The proposing fuzzy topic model work flow.

## II. Related Works

When it comes to the topic modeling, it is a norm to see using latent Dirichlet allocation (Blei, Ng et al. 2003) to get the keywords which contribute most to the same topic. However, it is not intuitive to set the wanted topic number in advance. The result is also posed for criticism because the output keyword is not related to the same topic. Even so, LDA still can be seen as a pioneer in this field and provided significant contribution since its introduction.

### A. *Autoencoder Models*

First of all, it starts from the basic encoder-decoder model, autoencoder (Hinton and Salakhutdinov 2006), which is introduced by GE Hinton in 2006. This work presents an unsupervised way to train a model. As presented in the paper, it embeds an image into a flat dimension vector, such as MNIST, and then prints it out in two dimensions map. This work shows the way to convert data into a meaningful matrix so we can use the matric to expand the application. Later, the paper of attention is all you need (Vaswani, Shazeer et al. 2017) shows an elated way to make a model to pay attention to an essential location in a matrix. The same model is used in image tasks as the following, (Anderson, He et al. 2018) (Pedersoli, Lucas et al. 2017) (Xu, Ba et al. 2015). Recently, there is a range of generative adversarial networks, (Radford, Metz et al. 2015) (Arjovsky, Chintala et al. 2017) (Mao, Li et al. 2017) which trains itself for growing a better image classification and image generating capability. In the field of natural language processing application, a trend has been provoked by ELMO (Peters, Neumann et al. 2018), BERT (Devlin, Chang et al. 2018), and GPT(Radford, Narasimhan et al. 2018). With the released pre-trained weight of BERT, people can easily manipulate the model to do a variety down streaming works, such as masked LM, next sentence prediction, sentiment analysis, and document classification. In order to fix the token limited in BERT and asymmetries of information between pre-training and fine-tuning, XLNet(Yang, Dai et al. 2019) is introduced to fix the flaws and limitation in BERT and ELMO. Because of the unlimited input sentence length and more reliable embedding as XLNet offering, we use XLNet embedding as our transformer model.

### B. *Dimension Reduction Techniques*

In reality, it is normal that each data has multiple features. When clustering each document based on the closet Euclidean distance, it is necessary to make dimension reduction toward each document embedding. In the begging, linear discriminant analysis (Mika, Ratsch et al. 1999) is the easiest way to downscale the data because of projecting the data based on a linear function. Further, principal component analysis (Wold, Esbensen et al. 1987) projects data based on the normal distribution or bell shape distribution. Then, it outputs the best axis to represent every data in the same dimension. However, as the models mentioned above either assume the data is linear separable or follow Gaussian distribution, the models usually have terrible performance with complex data distribution. Nevertheless, t-distribution stochastic neighbour embedding (Maaten and Hinton 2008), t-SNE, performs well, uncommonly projecting data into two-dimension.

### C. *Clustering Analysis*

Clustering is an intuitive way to group similar data. Nowadays, there are many clustering models. Furthermore, each model has its strength to gather different type of data. K-means (Lloyd 1982) can group dataset into k groups with setting the k in advance. On the other hand, DBSCAN (Kriegel, Kröger et al. 2011) cluster data based on the density and also considers the un-reachable data point as noise. Nevertheless, it might accidentally view essential data as a noise so HDBSCAN (McInnes, Healy et al. 2017), which is also a hybrid clustering model with density and hierarchical based mechanism, can prevent the problem and reduce the possible of overfitting problem by setting the spanning tree.

Further, the above clustering models are still hard clustering. In the real world, the criteria are not necessarily binary. Therefore, a soft clustering, such as fuzzy c-means (Bezdek, Ehrlich et al. 1984) and HDBSCAN soft clustering, show a better path to indicate the belonging value toward each cluster.

## III. Fuzzy Topic Modeling

### A. *Document Embedding Using XLNet*

As this work directly utilizes the XLNet pre-trained weight as our transformer model, the embedding algorithm can be shown as:

$$E = XLNet(document) \tag{1}$$

where $E$ is the embedding of the document and $E \in \mathbb{R}^{1\times 2048}$. The reason we get the $\mathbb{R}^{1\times 2048}$ is we use pooling of mean to average every sentence embedding by column. In the different document, we might get 123 x 2048 matrix, which means there are 123 sentences in the document. Moreover, we extract the last hidden layer of XLNet. Thus, we will get 2048 dimension of embedding, because of 2048 neurons in that layer.

### B. *t-SNE for Dimension Reduction*

Directly process a 2048-dimension dataset is costing a lot of computing power. Also, it is hard to understand the distance in such a high dimensional space. To avoid the issue and prepare for the following clustering step, this work used t-SNE to scale the dataset to n x 2. In the meantime, because t-SNE can measure the distance distribution in advance, the data can be moved based on their similarity in t distribution. The equation can be seen in the following:

$$RE = tSNE(E) \tag{2}$$

where $RE$ is the projecting dimension in $\mathbb{R}^{1\times 2}$, and $E$ is the embedding of the document in $\mathbb{R}^{1\times 2048}$.

### C. *Soft Clustering in HDBSCAN*

In our clustering section, we use HDBSCAN soft clustering to group similar embedding in two dimensions. The membership is combined two membership mechanisms, distance-based and outlier-based membership. The distance-based membership is utilizing the mutual reachability distance:

$$d_{mreach-k}(x_1, x_2) = \max\{core_k(x_1), core_k(x_2), core_k(x_1, x_2)\} \tag{3}$$

where $d(x_1, x_2)$ is the original distance between point $x_1$ and point $x_2$, $core_k(x_1)$ means parameter of k for a point $x_1$ and denote as a core. In order to convert the distance into probability, the distance can be show as:

$$\lambda = \frac{1}{d} \tag{4}$$

where d is the distance as the former equation.

Then, we use softmax to scale each $\lambda$ towards belonging cluster.

On the other hand, a modified global-local outlier score from hierarchies, GLOSH, algorithm is also applied as another distance measurement in HDBSCAN soft clustering. The GLOSH is as:

$$GLOSH(x_i) = \frac{\lambda_{max}(x_i) - \lambda(x_i)}{\lambda_{max}(x_i)} \tag{5}$$

The further detail of GLOSH algorithm can be seen in (Campello, Moulavi et al. 2015). Similarly, the softmax is used to scale the GLOSH score.

If we denote every projecting document, $RE$, as the $x_i$, the middle way to combine both measurements into:

$$U_i = [softmax(\lambda_i) \times softmax(glosh(x_i))] \tag{6}$$

where $i$ denotes to the $i$-th point.

For now, we only receive a conditional probability in a vector. Therefore, by using the following conversion, we can easily get the true membership vector:

$$P(x \in C_i | \exists_j : x \in C_j) \times P(\exists_j : x \in C_j) = P(x \in C_i, \exists_j : x \in C_j) \tag{7}$$

where $C_i$ denotes as the $i$-th cluster, $C_j$ denotes as the $j$-th cluster.

After receiving the membership vector, the model can finally pick up the representative point by sorting the vector value from the largest as the heart or the most representative data.

### D. *The Processing of Fuzzy Topic Modeling*

Lastly, this work is applied to all the techniques above. Before showing the clear workflow of the proposed model, this work does two-step topic model in practical application, which means doing one document embedding, two times dimensional reduction and two times clustering. The reason is that this process model can take a detailed insight into one group. Another reason is one step topic model ends up getting a relative general profile. In order to explore more, projecting the document embedding and clustering based on the last clustering result are used. Thus, the overall proposed workflow is as:

1. Document embedding from XLNet.

**First phase:**

2. Dimensional reduction to two dimensions from t-SNE.
3. Find the best minimal cluster size and clustering form HDBSCAN soft clustering.
4. Pick up the point whose membership function is not equal to zero, which might be a noise.

**Second phase:**

5. Based on the previous result, non-zero membership value data, to re-project the document embedding from t-SNE.
6. Cluster the projecting vector form HDBSCAN soft clustering based on the minimal cluster size in 5.
7. Sort the membership value based from top-to-down and pick the top 5 largest membership value as our representative.

## IV. Application and Simulation Results

In this part, we reveal our practical result by using the fuzzy topic model. The two phases will be shown respectively. What we have used in the experiment are articles that are related to AR and VR market news. All the data is scraped from multiple open media domains during November 2019, which means the data is pre prepared. The data can be seen in Figure 2 with 3783 articles in total. However, it is believed the work in this paper can be easily implemented in other genre articles as LDA. In this section, the detail process and performance are presented. Then the result comparison between proposed work with LDA in the same dataset is shown in part C.

| 1 | id | title | raw_text | domain | url |
|---|---|---|---|---|---|
| 2 | 2509616 | New app for marathon runners | New York Road Runners in colla | appdevelope | https://appd |
| 3 | 2509656 | VR Game Review: OhShape Has Satisfying Rhythm & | OhShape , previously called Oı | arpost.co | https://arpo: |
| 4 | 2510923 | Virtual Reality Learning at Penrhos College | Sign In Get more than ever out | anz.business | https://anz.b |
| 5 | 2513231 | November 2019 Special Edition: 2019 Robotics Handb | DHL on state warehouse roboti | designworld | https://www |
| 6 | 2513614 | MG Motor Opens First Digital Showroom In Bangalor | MG Motor India has opened a | drivespark.cc | https://www |
| 7 | 2517165 | Augmented Reality assists surgeons in the operating | Artificial intelligence is taking c | innovationor | https://innov |
| 8 | 2520395 | Gameboard-1 Is a Digital Game Console Built for the | Gameboard-1, a project curren | nerdist.com | https://nerdi |
| 9 | 2522150 | PSVR Game Golem Delayed One More Week Right Be | If you,Äôve been excited for Pla | playstationli | https://www |
| 10 | 2524345 | Facebook Was Reportedly Looking to Buy FitBit, and N | One of the biggest tech industr | socialmediat | https://www |
| 11 | 2524442 | November's Most Important Air Jordan Release Dates | Release Date: 11/01/2019 Styl | solecollector | https://solec |
| 12 | 2524890 | Games Watch: the 5 best new games coming out in N | Even during the slowest month | stuff.tv | https://www |
| 13 | 2525161 | Google Pixel 4 Review ,Äì Outstanding, but for the Ba | A stellar phone let down only b | tech.co | https://tech. |
| 14 | 2527495 | Five Nights At Freddy's VR: Help Wanted Is Coming T | Popular horror game Five Night | uploadvr.con | https://uploa |
| 15 | 2527501 | PSVR's Golem Delayed One Last Time (Maybe, Hopef | I can,Äôt believe I,Äôm writing | uploadvr.con | https://uploa |
| 16 | 2527505 | Doctor Who VR Arcade Game Spinning Out Of Home | Want to jump into the world of | uploadvr.con | https://uploa |
| 17 | 2527506 | Layers Of Fear Is Finally Coming To Oculus Rift and H | The original Layers of Fear gam | uploadvr.con | https://uploa |
| 18 | 2527507 | Oculus Link Beta: What To Expect From Quest's PC VF | Oculus Link addresses one of th | uploadvr.con | https://uploa |
| 19 | 2527512 | Competition: Win An Oculus Quest Headset With Synt | In a dancing mood? Want an O | uploadvr.con | https://uploa |
| 20 | 2527523 | Spaces Dead, Tilt Five Funded And Win An Oculus Qu | It,Äôs been a week of trials anc | uploadvr.con | https://uploa |

Figure 2. Experiment data in November, 2019.

### A. First Phase

As the previous section mentioned about the document embedding from XLNet, each document embedding has the shape of (1, 2048). Then, we project each 2048-dimensional data into two-dimensional space as Figure 3.

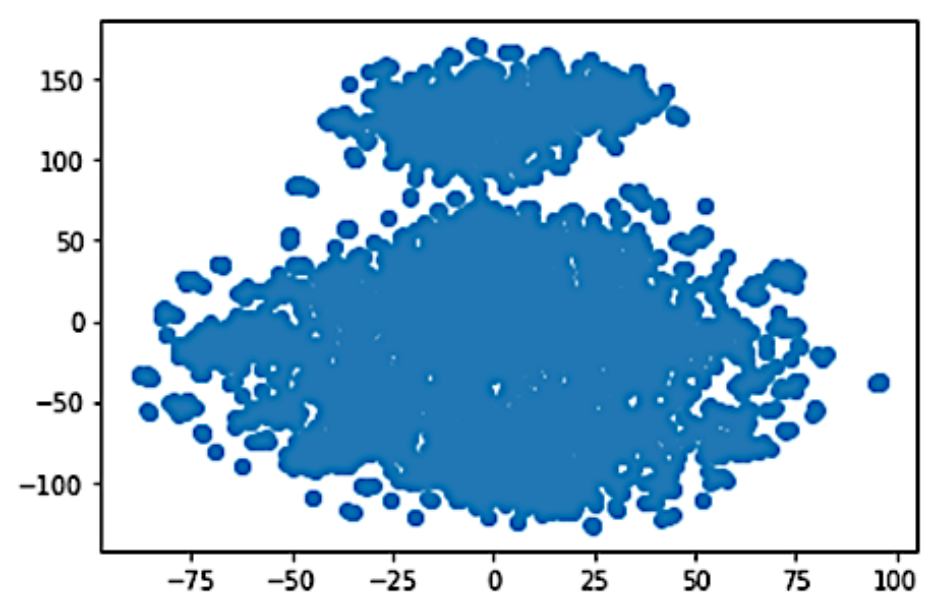

Figure 3. Document distribution after t-SNE projecting in two dimensions.

Then, in order to find the best minimal cluster size and prevent a manual setting, the best persistence score from the grid search in the overall cluster is as Figure 4.

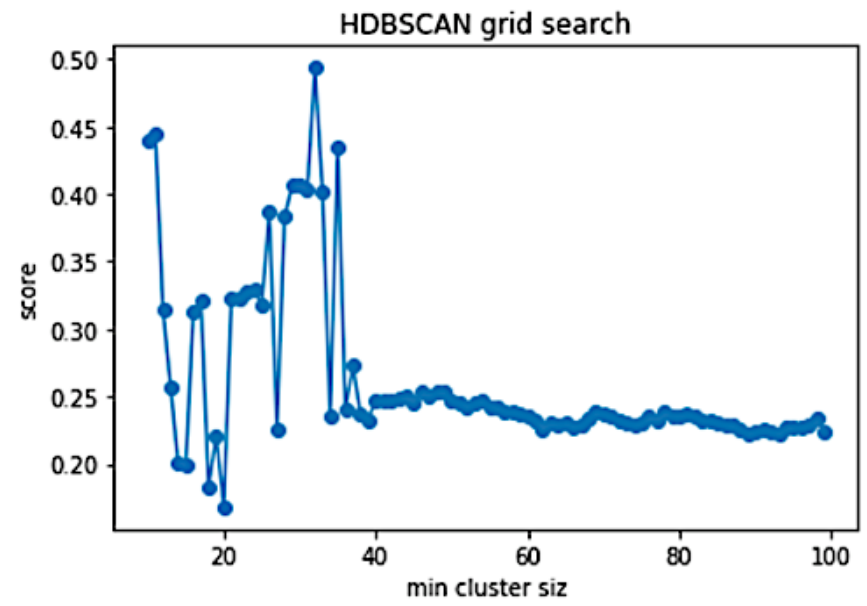

Figure 4. The grid search result. (After the grid search, we find the best minimal cluster size is 32 in this dataset.)

Therefore, based on the gird search result, the HDBSCAN minimal cluster size is 32. Moreover, the result of using a membership value to pick up the representative data is as Figure 5 and Figure 6. The colour saturability represents the membership vector value.

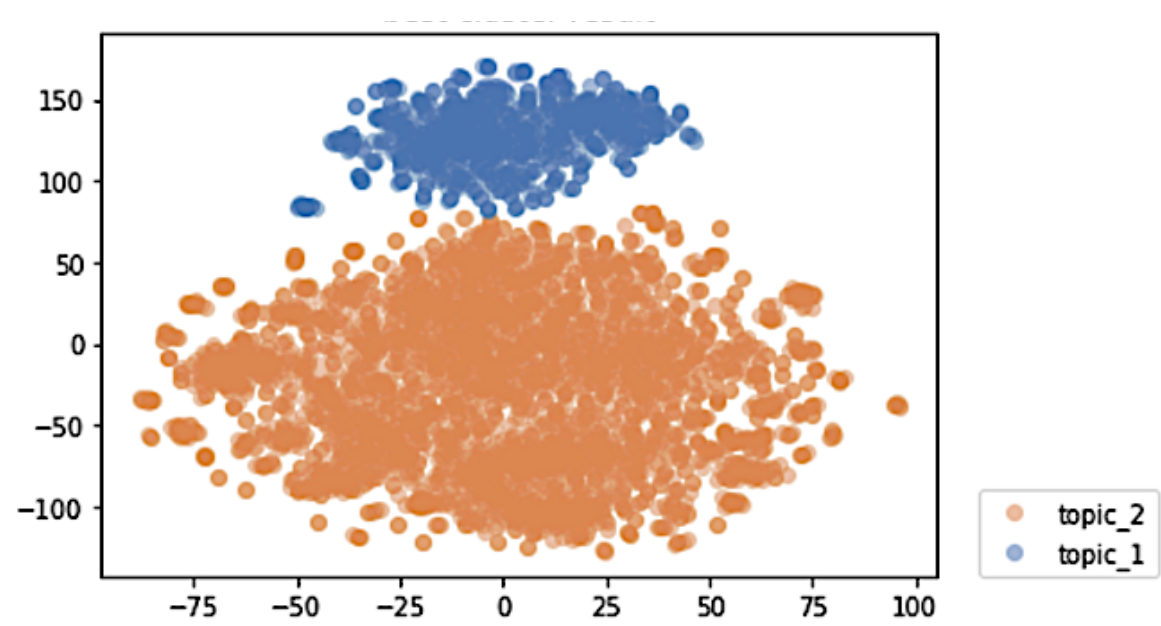

Figure 5. The soft clustering result of the phase 1. (We can see the two groups can be separated clearly. However, this is the general result.)

| 1 | topic_label | article_id | title | cluster_m |
|---|---|---|---|---|
| 2 | topic_2 | 2670034 | Apple reportedly plans 2022 release for first AR headset, followed by AR glasses in 2023 | 1 |
| 3 | topic_2 | 2670509 | Minecraft Earth Lands in the US—Let the Block Party Begin | 1 |
| 4 | topic_2 | 2677706 | The Louvre Re-created the Mona Lisa in 3D in Painstaking Detail | 1 |
| 5 | topic_2 | 2669979 | Fordham University business students have a new tool to prepare them for boardrooms: Virtual reali | 1 |
| 6 | topic_2 | 2674750 | 5G brings "an age of magic" | 1 |
| 7 | topic_1 | 2670442 | AR/VR Lens market augmented expansion to be registered until 2024 market players are Luxexcel C | 1 |
| 8 | topic_1 | 2672056 | VR for Medical Market 2019 Strategic Assessments – Psious, zSpace, MindMaze, CanquerMobile, | 1 |
| 9 | topic_1 | 2672286 | VR glove Industry 2019-2024l Market Size, Trends, Regional Outlook, Opportunities, Demands and | 1 |
| 10 | topic_1 | 2672300 | VR Gaming Market : Worldwide Industry Analysis and New Market Opportunities Explored; 2025 - | 1 |
| 11 | topic_1 | 2672331 | Global Vr Gloves Market 2019 Growth and Share Analysis By Top players, Application, and Types | 1 |

Figure 6. The representative data based on the membership value. (We can easily tell the topic 1 is related to VR but the topic 2's content is not specific.)

### B. Second Phase

In order to extract a more specific result, as taking insight is also a vital goal, it is necessary to do a phase II fuzzy topic modeling. As a result, the document embedding of every non-zero vector data is first re-projected based on the previous clustering result. Second, to get a top-5 list of representative articles, minimal cluster size is set as 5 in this phase. The fuzzy topic modeling towards the result of each topic can be shown in Figure 7 and Figure 8. Again, the colour saturability represents the membership vector value.

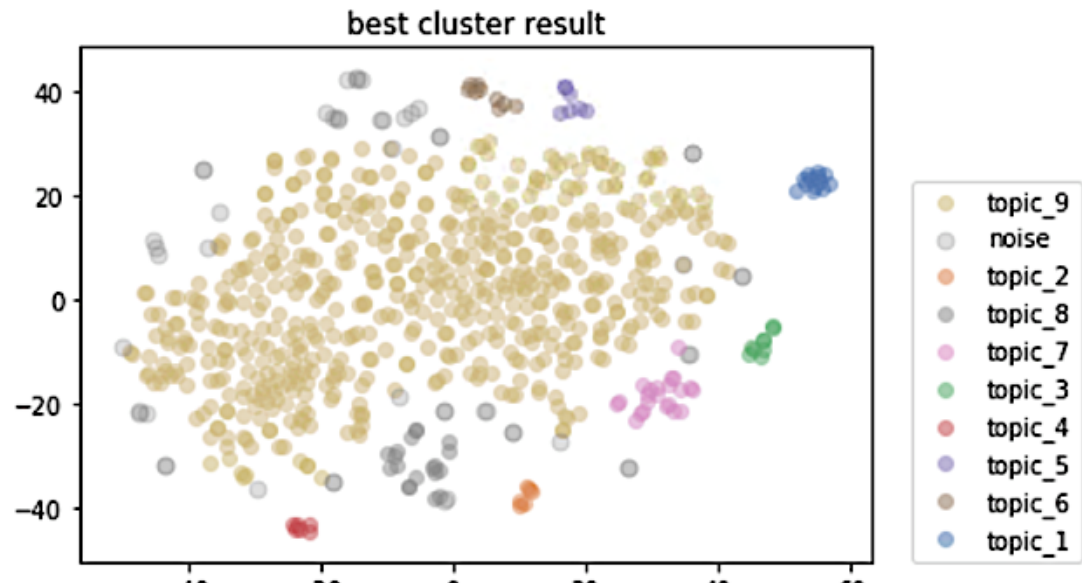

Figure 7. The topic modeling of topic 1 - phase 2. (Now, we can easily extract minor topic and exclude the noise in the topic 1 group. Except noise, we get nine topics.)

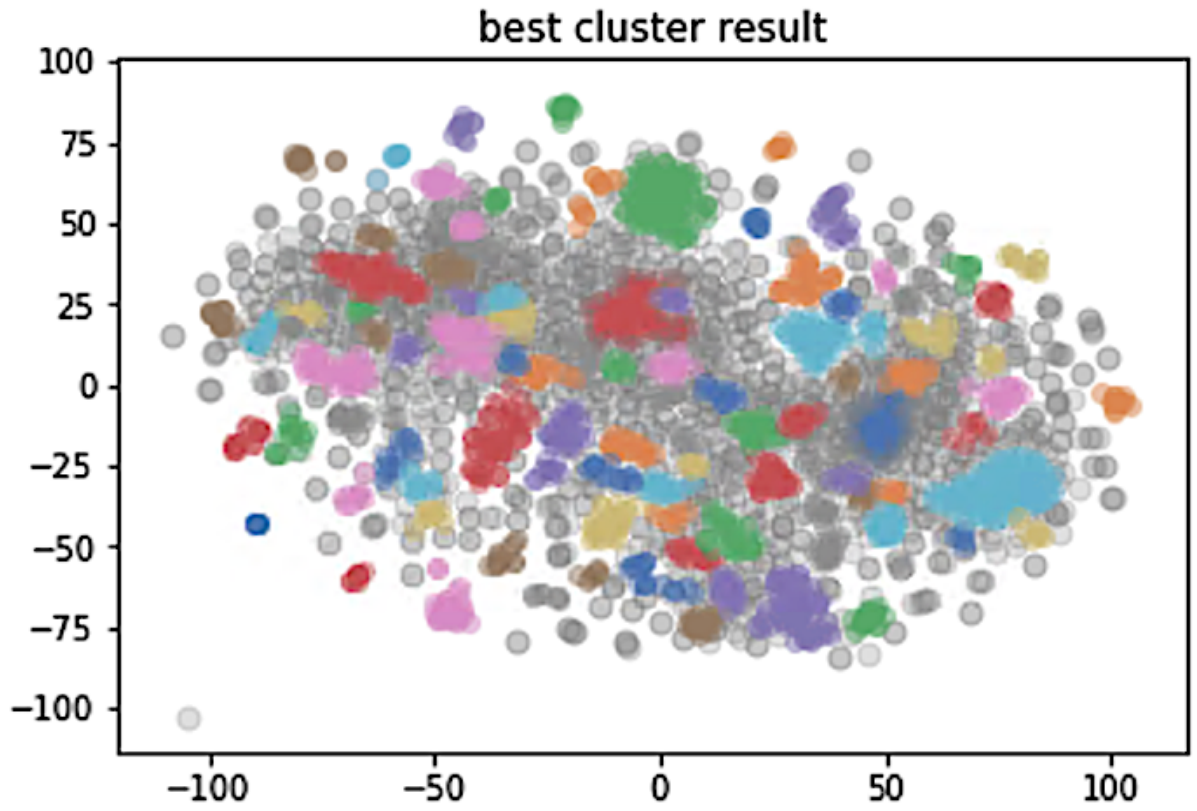

Figure 8. The topic modeling of the topic 2 - phase 2. (More minor topics can be seen in the group. Also, some not specific or noisy article can be removed by using membership vectors. As the number of minor topics is too messy to show, the index of each minor topic label is not revealed.)

Finally, the result of fuzzy topic model after two phases modeling is under below:

| 1 | topic_label | article_id | title | raw_text | cluster_m |
|---|---|---|---|---|---|
| 2 | topic_43 | 2660091 | Report: Apple Targeting 2022 For Oculus Quest Competitor, | Apple may | 1 |
| 3 | topic_43 | 2674210 | Apple Partners with Valve to Develop its Rumoured AR Heac | Apple has | 0.8906 |
| 4 | topic_43 | 2671340 | Apple revealed its Augmented Reality helmet schedule for 20: | There were | 1 |
| 5 | topic_43 | 2673381 | Valve Corporation And Apple Team To Create An AR Heads | The future | 0.823315 |
| 6 | topic_43 | 2673782 | Valve and Apple Working on AR Gaming Headset? | Apple is sa | 0.895258 |

Figure 9. The result of the topic 43 in fuzzy topic model. (The topic is mainly discussing apple's VR.)

| 1 | topic_label | article_id | title | raw_text | cluster_m |
|---|---|---|---|---|---|
| 8 | topic_29 | 2673477 | Microsoft' s HoloLens 2 is ready for work | After a sen | 0.931876 |
| 9 | topic_29 | 2671341 | Microsoft' s HoloLens 2 Now Available for $3,500 | Microsoft : | 1 |
| 10 | topic_29 | 2673338 | Varjo VR-2 virtual reality headset delivers "human-eye resolu | If you're p | 0.706308 |
| 11 | topic_29 | 2673215 | Microsoft just launched the $3,500 HoloLens 2 augmented re: | Announced | 1 |

Figure 10. The result of the topic 29 in fuzzy topic model. (It is talking about Microsoft's HoloLens2.)

| 1 | topic_label | article_id | title | raw_text | cluster_m |
|---|---|---|---|---|---|
| 17 | topic_35 | 2673057 | Adobe Aero turns Photoshop layers into interactive AR exper: | Adobe is re | 1 |
| 18 | topic_35 | 2674186 | Adobe's Next-Generation Apps Promise to Make Collaboratic | Adobe firs | 1 |
| 19 | topic_35 | 2673609 | Photoshop for iPad had finally arrived, Adobe announces Illu | CNET tam | 1 |
| 20 | topic_35 | 2674540 | Adobe's advanced AI editing tools graduate to Creative Cloud | Adobe has | 1 |
| 21 | topic_35 | 2673764 | Adobe Aero launch puts the power of augmented reality in th | First previe | 0.92108 |

Figure 11. The result of the topic 35 in fuzzy topic model. (The topic can easily be observed, talking about Adobe.)

Another important note, the proposed process only embeds the text of articles instead of directly embedding the title. Therefore, it is quite motivating to see the method able to group the same title. Most importantly, the column of 'cluster_m' represents the softmax membership vector. The value is related to the centre of the topic group.

As the outputted topic number of the proposed method is in great number, the sample of the result in top 5 topic is shown in the following Figure 12:

| 1 | topic_label | title |
|---|---|---|
| 2 | topic_43 | Report: Apple Targeting 2022 For Oculus Quest Competitor, 2023 For AR Glasses - Uploa |
| 3 | topic_43 | Apple Partners with Valve to Develop its Rumoured AR Headset \| NewsGram |
| 4 | topic_43 | Apple revealed its Augmented Reality helmet schedule for 2022 in Steve Jobs Theater - O |
| 5 | topic_43 | Valve Corporation And Apple Team To Create An AR Headset |
| 6 | topic_43 | Valve and Apple Working on AR Gaming Headset? |
| 7 | topic_29 | Microsoft HoloLens 2 offers trippy, eyeball-tracking augmented reality |
| 8 | topic_29 | Microsoft's HoloLens 2 is ready for work |
| 9 | topic_29 | Microsoft's HoloLens 2 Now Available for $3,500 |
| 10 | topic_29 | Varjo VR-2 virtual reality headset delivers "human-eye resolution" |
| 11 | topic_29 | Microsoft just launched the $3,500 HoloLens 2 augmented reality headset |
| 12 | topic_84 | Apple Said to Prep AR Headset for 2022 |
| 13 | topic_84 | The Makers Of James Dean CGI Movie Unphased By Backlash, Expanding Company |
| 14 | topic_84 | Sony embraces the absurd in sci-fi spot for PlayStation VR |
| 15 | topic_84 | Looking Glass Factory Unveils 8K Holographic Display |
| 16 | topic_84 | Team Behind Digital James Dean Forms New Company to Resurrect Other Legends (EXCLU |
| 17 | topic_35 | Adobe Aero turns Photoshop layers into interactive AR experiences |
| 18 | topic_35 | Adobe's Next-Generation Apps Promise to Make Collaboration Simpler |
| 19 | topic_35 | Photoshop for iPad had finally arrived, Adobe announces Illustrator is coming next |
| 20 | topic_35 | Adobe's advanced AI editing tools graduate to Creative Cloud apps |
| 21 | topic_35 | Adobe Aero launch puts the power of augmented reality in the hands of artists - SiliconAI |
| 22 | topic_34 | Researchers Create 3D Images That Can Play Sound, React to Touch |
| 23 | topic_34 | dearVR PRO gets a major update on its multichannel spatialisation - MusicTech |
| 24 | topic_34 | Researchers Develop Method to Boost Contrast in VR Headsets by Lying to Your Eyes |
| 25 | topic_34 | Calling Princess Leia: How the out-of-this-galaxy Star Wars hologram just grew to becom |
| 26 | topic_34 | Apple patents tangibility visualization of virtual objects within headsets |
| 27 | topic_81 | Golem Is Out On PlayStation VR From Highwire, Finally Releases After Years Of Delays Pus |
| 28 | topic_81 | Pistol Whip review: The year's freshest VR game—and oh-so close to greatness - The Medi |
| 29 | topic_81 | Stormland Review: VR's Slickest Shooter Yet (But Not Without Issue) |
| 30 | topic_81 | Star Trek: Discovery Away Mission Review – Engaging Trekkies At SandboxVR |
| 31 | topic_81 | Transformers VR Coming To Arcades From Minority Media |

Figure 12. The sample of proposed method output.

## C. *Comparison with LDA*

As the same argument in the previous section, although LDA has shown its merit in using conditional probability on a word to show the possible topic among text, setting topic number in advance can be viewed as an unnatural process from the aspect of observing market information. As for how do people know how many topics already exist in the presenting market news? In order to avoid human factor among the comparison, the topic number was set by applying a grid search. The number of topics was determined by the coherence value, which is introduced from (Röder, Both et al. 2015). The calculating of coherence value follows the pipeline of segmentation, probability estimation, confirmation measure, and aggregation. Among the steps, we removed stop words in advance in the article content for segmentation and build the corpus for probability estimation. Lastly, the topic number range was set from 2 to 10 in 2 steps. The highest coherence value of topic number then is chosen as the final topic number for the following comparison.

Based on the description of the comparison experiment, the coherence value of each topic number is presented in the following Table 1 and Figure 13:

Table 1. The grid search result in LDA.

| Topic Number | Coherence Value |
|---|---|
| 2 | 0.555 |
| 4 | 0.5595 |
| 6 | 0.5877 |
| 8 | 0.5307 |
| 10 | 0.5496 |

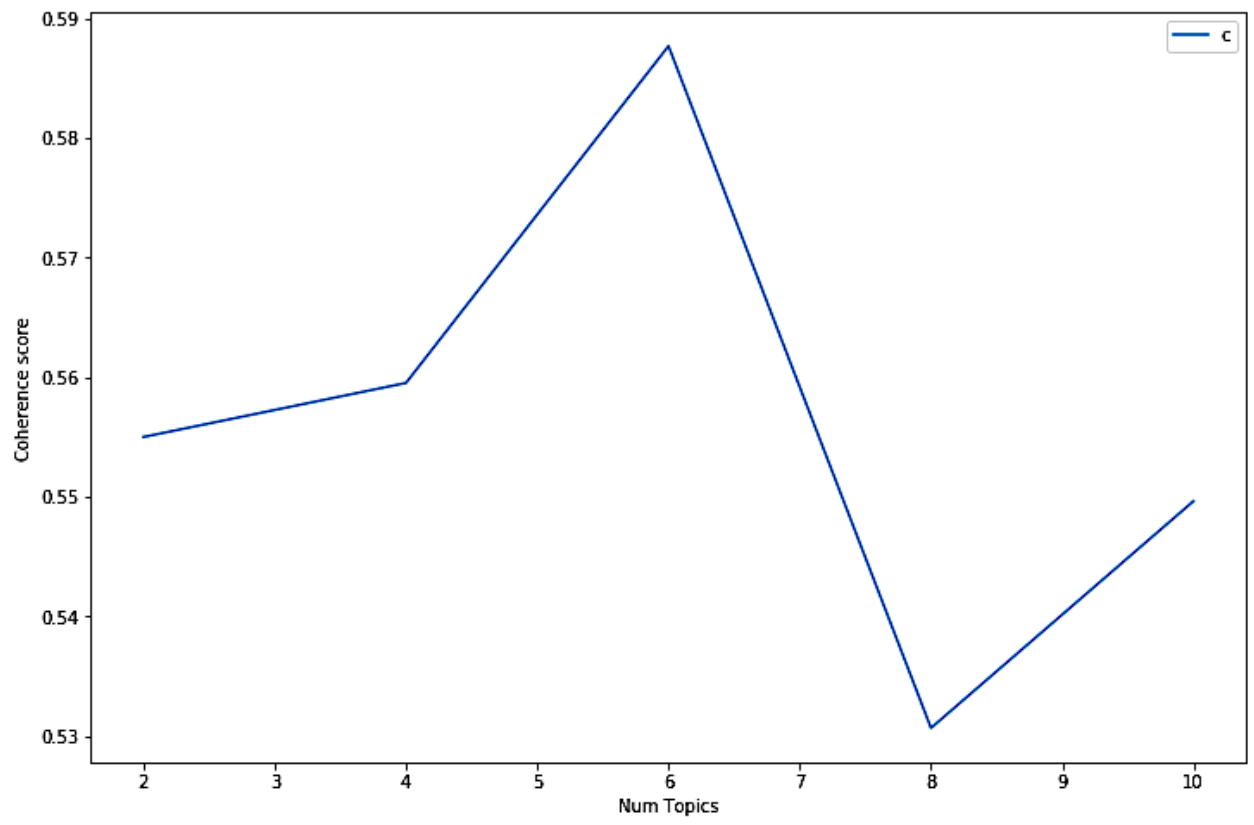

Figure 13. The grid search result in LDA.

From the hyperparameter adjustment result, the topic number in LDA was set to 6 because the coherence value of topic number 6 was the highest. Therefore, the represented articles in each topic are revealed in the following:

| 1 | Topic_Num | title | Keywords |
|---|---|---|---|
| 2 | 0 | The Best Smart Notebooks of 2019 \| Digital Trends | apple, app, device, camera |
| 3 | 0 | How a mini projector can transform your movie nights | apple, app, device, camera |
| 4 | 0 | Samsung Galaxy S10 range gets even more great Note 10 features \| Trusted Reviews | apple, app, device, camera |
| 5 | 0 | Spotify's Magic Leap App Allows You to Pin Music in Your Home | apple, app, device, camera |
| 6 | 0 | Snapchat adds Time Machine lens for aging and de-aging of your face | apple, app, device, camera |

Figure 14. The result of topic 1. (The topic is related to Apple or app based on the keyword)

Compared the topic related to Apple with the result of the proposed method in topic 43, although LDA computed the topic based on the keywords, including apple, app, and device, the represented articles were not as good as the result of topic 43. On the other side, LDA still offered a good result in the topic of PS4 as below:

| 1 | Topic_Num | title | Keywords |
|---|---|---|---|
| 27 | 5 | GameStop Black Friday 2019 Deals Revealed: PS4 Pro, Xbox One X Bundles, Cheap Game: | game, deal, console, black_friday, playstation, xbox, bunc |
| 28 | 5 | GameStop Black Friday Deals 2019: Best Deals On Nintendo Switch, PS4, Xbox One, And I | game, deal, console, black_friday, playstation, xbox, bunc |
| 29 | 5 | PS4 and PS4 Pro Black Friday deals - what to expect in 2019 | game, deal, console, black_friday, playstation, xbox, bunc |
| 30 | 5 | PS4 and PS4 Pro Black Friday deals - what to expect in 2019 | game, deal, console, black_friday, playstation, xbox, bunc |
| 31 | 5 | PS4 and PS4 Pro Black Friday deals - live offers and what to expect in 2019 | game, deal, console, black_friday, playstation, xbox, bunc |

Figure 15. The result of Topic 5 in LDA. (The topic is related to PS4 based on the keyword of game. deal, and etc.)

As Figure 15 shown, LDA produced a decent result related to PS4, but another question was raised towards the topic number. Compared the result with Fuzzy topic model in following Figure 16 to Figure 18:

| 1 | topic_label | title |
|---|---|---|
| 65 | topic_45 | Kohl's Black Friday 2019 Sales: A look at their gaming section |
| 66 | topic_45 | PlayStation Store Free Games for the Week of Nov. 5 (DLC, Themes & More) |
| 67 | topic_45 | Kohl's Black Friday 2019: PlayStation, Xbox and Nintendo gaming deals revealed |
| 68 | topic_45 | Black Friday 2019: Kohl's Deals For PS4, Xbox One, and Switch With Great Bonuses |
| 69 | topic_45 | Black Friday 2019 deals at Kohl's: What you need to know |

Figure 16. The result of topic 45 in Fuzzy topic model. (The topic is related to PS4.)

| 1 | topic_label | title |
|---|---|---|
| 121 | topic_33 | Mini Motor Racing X Brings Micro Machines Style Action to PS4 and PSVR Next Month |
| 122 | topic_33 | New PSVR Ad Takes You Hostage and Welcomes an Alien Invasion |
| 123 | topic_33 | Pistol Whip's Kannibalen Records Tracklist Revealed, Listen On Spotify |

Figure 17. The result of topic 33 in Fuzzy topic model. (The topic is related to PS4.)

| 1 | topic_label | title |
|---|---|---|
| 101 | topic_24 | Doctor Who VR game Doctor Who: The Edge of Time out now |
| 102 | topic_24 | Dr. Who: The Edge of Time is out today on PC and PS4 - Game Freaks 365 |

Figure 18. The result of topic 24 in Fuzzy topic model. (The topic is related to PS4.)

Figure 16 to Figure 18 has shown the topic related to PS4 more than LDA result. The proposed model not only outputs the PS4 news related to black Friday but also reveals the VR game in PS4. This result shows better coverage among different spectrum topic toward the same big topic, which LDA cannot offer. The reason could be contributed to using contextual meaning from word embedding instead of only calculating conditional probability in LDA. Also, with the merit of unsupervised clustering in HDBSCAN, the several iterations and only setting minimum cluster size helps to exhibit continuously close data point. Last but not least, assigning the fuzzy membership value to data point makes the represented data more meaningful. In order to compare the result between the proposed method and LDA, the output of LDA is in Figure 19.

| 1 | Topic_Num | title | Keywords |
|---|---|---|---|
| 2 | 0 | The Best Smart Notebooks of 2019 \| Digital Trends | apple, app, device, camera, headset, video, user, feature |
| 3 | 0 | How a mini projector can transform your movie nights | apple, app, device, camera, headset, video, user, feature |
| 4 | 0 | Samsung Galaxy S10 range gets even more great Note 10 features \| Trusted Reviews | apple, app, device, camera, headset, video, user, feature |
| 5 | 0 | Spotify's Magic Leap App Allows You to Pin Music in Your Home | apple, app, device, camera, headset, video, user, feature |
| 6 | 0 | Snapchat adds Time Machine lens for aging and de-aging of your face | apple, app, device, camera, headset, video, user, feature |
| 7 | 1 | Stormland is an Oculus exclusive that pushes the boundaries of VR | game, vr, headset, player, reality, valve, experience, ocul |
| 8 | 1 | Golem Review - Innovation for the Worse | game, vr, headset, player, reality, valve, experience, ocul |
| 9 | 1 | Valve finally announces another Half-Life game \| TheINQUIRER | game, vr, headset, player, reality, valve, experience, ocul |
| 10 | 1 | Half-Life: Alyx is officially coming March 2020, and here‚Äôs your first look | game, vr, headset, player, reality, valve, experience, ocul |
| 11 | 1 | New Half-Life Game Answers Many, Many Prayers - Comic Years | game, vr, headset, player, reality, valve, experience, ocul |
| 12 | 2 | AI and 5G: Entering a new world of data | technology, reality, experience, customer, product, datum |
| 13 | 2 | A Virtual Hospital, A Nav App For The Blind, An AR Microscope: Israel Showcases Digital I | technology, reality, experience, customer, product, datum |
| 14 | 2 | Improving medicines adherence with VR - | technology, reality, experience, customer, product, datum |
| 15 | 2 | Building A More Collaborative AR Platform | technology, reality, experience, customer, product, datum |
| 16 | 2 | In Japan, Advanced Technology Helps Seniors To Live Better \| Naija News | technology, reality, experience, customer, product, datum |
| 17 | 3 | Augmented Reality (AR) and Virtual Reality (VR) in Healthcare Market Astonishing Growt | market, report, reality, analysis, research, growth, vr, app |
| 18 | 3 | Augmented Reality And Virtual Reality Component Market Development History, Current | market, report, reality, analysis, research, growth, vr, app |
| 19 | 3 | Exclusive Report on Augmented Reality (AR) and Virtual Reality (VR) in Healthcare Marke | market, report, reality, analysis, research, growth, vr, app |
| 20 | 3 | Demandable Report: Augmented Reality (AR) and Virtual Reality (VR) in Healthcare Mark | market, report, reality, analysis, research, growth, vr, app |
| 21 | 3 | Augmented Reality (AR) and Virtual Reality (VR) in Healthcare Market Comprehensive an | market, report, reality, analysis, research, growth, vr, app |
| 22 | 4 | Here‚Äôs where you can watch Japanese movies on a big screen in Chennai | experience, reality, event, vr, film, art, story, project, vide |
| 23 | 4 | Cineplex announces new entertainment destination Junxion \| blooloop | experience, reality, event, vr, film, art, story, project, vide |
| 24 | 4 | Jet off to space as a tourist from US, virtually of course | experience, reality, event, vr, film, art, story, project, vide |
| 25 | 4 | Hersheypark announces new virtual reality attraction | experience, reality, event, vr, film, art, story, project, vide |
| 26 | 4 | BlastOne Features Robotic Blasting, Coating Technology | experience, reality, event, vr, film, art, story, project, vide |
| 27 | 5 | GameStop Black Friday 2019 Deals Revealed: PS4 Pro, Xbox One X Bundles, Cheap Game: | game, deal, console, black_friday, playstation, xbox, bunc |
| 28 | 5 | GameStop Black Friday Deals 2019: Best Deals On Nintendo Switch, PS4, Xbox One, And I | game, deal, console, black_friday, playstation, xbox, bunc |
| 29 | 5 | PS4 and PS4 Pro Black Friday deals - what to expect in 2019 | game, deal, console, black_friday, playstation, xbox, bunc |
| 30 | 5 | PS4 and PS4 Pro Black Friday deals - what to expect in 2019 | game, deal, console, black_friday, playstation, xbox, bunc |
| 31 | 5 | PS4 and PS4 Pro Black Friday deals - live offers and what to expect in 2019 | game, deal, console, black_friday, playstation, xbox, bunc |

Figure 19. The output of LDA.

## V. CONCLUSION

Given the exciting result of our method, it is recommended to replace the typical LDA by combining document embedding and HDBSACN soft clustering since, as the experiments have shown, the method picks the topic number automatically and offers a more nature and precise result. Even in the first phase of topic modeling, the chosen topic already reveals a particular topic. Conversely, in section IV experiment comparison with LDA, the topic analysis result from LDA is limited because of setting topic number in advance. Also, how to pick up represented articles, with the blessing by the membership vector, experiments show the result of the proposed method is much meaningful and intuitive as the process of setting topic number manually and possible of showing general articles are avoided in the suggested process. All in all, the proposed method is quite suitable for people who want to use the NLP model to monitor market trending news without involving human factor. Most importantly, with the coming era of using pre-trained weight in state-of-the-art transformers, applying the proposed method in real-world tasks can easily transcend the traditional method and provide a precise but also sensitive information towards small subset. The above character is an essential feature in the rapidly changing market nowadays.

However, there are some flaws in our method, including data noise. Also, the clustering result is be affected by the projecting method. Thus, although using XLNet document embedding is a state-of-the-art method, the work of how to project the embedding to cluster is still an on-going task. It is doable to train a transformer from scratch to decode the transformer, BERT and XLNet, embedding so we can use the autoencoder to project data to low dimension.

## ACKNOWLEDGEMENTS

This research is partially supported by the Ministry of Science and Technology research grant in Taiwan (MOST 108-2218-E-009-041 -).